\documentclass[10pt,twocolumn,letterpaper]{article}
\usepackage[T1]{fontenc}
\usepackage{times}
\usepackage[textwidth=6.875in,textheight=8.875in,top=0.75in,columnsep=0.3125in]{geometry}
\usepackage[dvipsnames]{xcolor}
\usepackage{graphicx,amsmath,amssymb,booktabs,array}
\usepackage[numbers,sort&compress]{natbib}
\usepackage[format=plain,labelformat=simple,labelsep=period,font=small,compatibility=false]{caption}
\usepackage[font=footnotesize,skip=3pt,subrefformat=parens]{subcaption}
\usepackage[shortlabels,inline]{enumitem}
\setlist[itemize]{noitemsep,leftmargin=*,topsep=0em}
\setlist[enumerate]{noitemsep,leftmargin=*,topsep=0em}
\usepackage[hyphens]{url}
\usepackage[breaklinks,colorlinks,allcolors=blue]{hyperref}
\hypersetup{
  pdftitle={ProbeShare: Probing a Few, Sharing Across Many for Fast World-Model Planning},
  pdfauthor={Zejian Kang, Xiantao Lin, Zekai Chen, Boyang Shen, Mingyu Ma, Shengwen Zheng, Chengtai Guo, Xiangru Huang},
  pdfkeywords={robot planning, latent world models, model predictive control, action-conditioned prediction, cross-entropy method}
}
\newcommand{\figuretype}{\rmfamily}
\title{ProbeShare: Probing a Few, Sharing Across Many\\for Fast World-Model Planning}
\author{%
  Zejian Kang\textsuperscript{1,2,*}\quad
  Xiantao Lin\textsuperscript{3,*}\quad
  Zekai Chen\textsuperscript{4}\quad
  Boyang Shen\textsuperscript{5}\\[4pt]
  Mingyu Ma\textsuperscript{6}\quad
  Shengwen Zheng\textsuperscript{7}\quad
  Chengtai Guo\textsuperscript{4}\quad
  Xiangru Huang\textsuperscript{2,\ensuremath{\dagger}}\\[8pt]
  {\small\textsuperscript{1}Zhejiang University\quad
  \textsuperscript{2}Westlake University}\\[2pt]
  {\small\textsuperscript{3}Southwest Jiaotong University\quad
  \textsuperscript{4}Beijing Institute of Technology}\\[2pt]
  {\small\textsuperscript{5}Huazhong University of Science and Technology}\\[2pt]
  {\small\textsuperscript{6}Northwest Polytechnical University\quad
  \textsuperscript{7}National University of Defense Technology}\\[6pt]
  {\small\textsuperscript{*}Equal contribution.\quad
  \textsuperscript{\ensuremath{\dagger}}Corresponding author:
  \href{mailto:huangxiangru@westlake.edu.cn}{huangxiangru@westlake.edu.cn}}
}
\date{}

\makeatletter
\providecommand{\@LN}[2]{}
\renewcommand{\section}{\@startsection{section}{1}{\z@}
  {-10pt plus -2pt minus -2pt}{7pt}{\normalfont\large\bfseries}}
\renewcommand{\subsection}{\@startsection{subsection}{2}{\z@}
  {-8pt plus -2pt minus -2pt}{5pt}{\normalfont\fontsize{11}{13}\selectfont\bfseries}}
\renewcommand{\subsubsection}{\@startsection{subsubsection}{3}{\z@}
  {-6pt plus -2pt minus -2pt}{3pt}{\normalfont\normalsize\bfseries}}
\renewcommand{\@maketitle}{%
  \newpage\null
  \begin{center}
    {\Large\bfseries\@title\par}
    \vskip 12pt
    {\large\lineskip .5em\begin{tabular}[t]{c}\@author\end{tabular}\par}
  \end{center}
  \vskip 6pt
\begin{minipage}{\textwidth}
\centering
\includegraphics[width=0.85\linewidth]{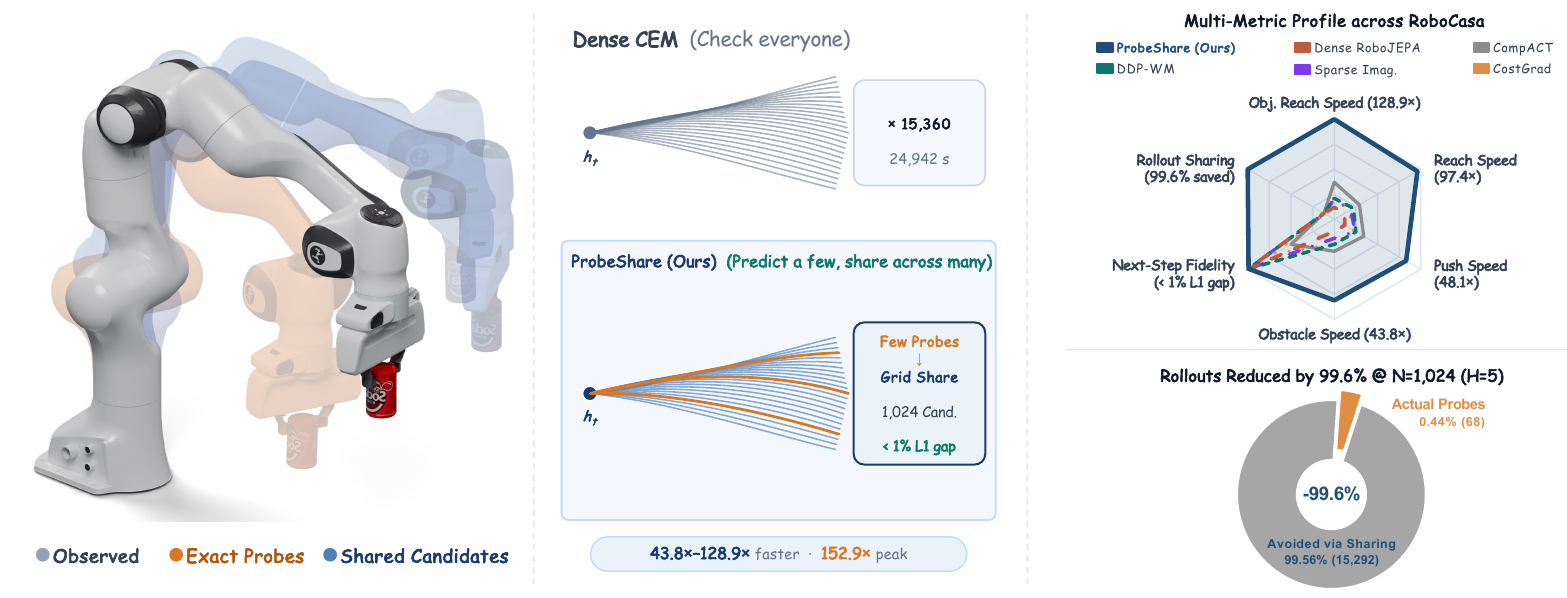}
\captionof{figure}{\textbf{Predict a few, share across many.} \textbf{Left} shows the observed robot state (gray), probe futures predicted by the frozen world model (orange), and reconstructed candidate futures (blue). \textbf{Middle} compares dense CEM's individual rollouts with ProbeShare's reuse of probe feature grids across 1,024 candidate sequences. \textbf{Right} compares planning speed and next-step feature fidelity across RoboCasa tasks. The ring shows a $99.6\%$ reduction in full rollouts for five-step searches with 1,024 candidates per round on A800. Peak fixed-input speedup reaches $152.9\times$ on Object Reach with the RTX A6000.}\label{fig:teaser}
\end{minipage}\par\vspace{4pt}
}
\makeatother
\renewenvironment{abstract}{%
  \centerline{\large\bfseries Abstract}\vspace{12pt}\noindent\itshape\ignorespaces
}{\par\vspace{12pt}}

\begin{document}
\maketitle
\begin{abstract}
\widowpenalty=10000
Visual planning with a frozen joint-embedding world model supports robotic reaching and object manipulation by evaluating candidate action sequences against an image goal before execution. Standard cross-entropy method (CEM) planners roll out every candidate independently through a billion-parameter predictor, even when nearby actions induce similar scene changes. This redundant computation grows with population size and search rounds, slowing closed-loop replanning. We observe that candidates within a planning step share an observation history and thus a common action-to-feature map. Based on this insight, we propose ProbeShare, which measures local responses with a few representative probes and reconstructs each candidate's feature grid through scene-conditioned, geometry-aware weighting. When promising candidates move beyond existing probe coverage, ProbeShare evaluates a new probe to rescore nearby candidates. Across four RoboCasa tasks, ProbeShare accelerates episode planning by \mbox{$43.8$--$128.9\times$} over dense RoboJEPA and at least $14.2\times$ over competitive baselines, with greater mean goal progress in a matched-time comparison.\par
\end{abstract}
\section{Introduction}
A robot reaching a target evaluates candidate motions with different directions and step sizes. A joint-embedding world model predicts the visual features at the end of each proposed action sequence~\cite{vjepa2,robojepa}. The cross-entropy method (CEM)~\cite{cemoriginal} ranks candidate sequences by their feature distance to the goal image and samples new sequences around the lowest-cost candidates. Selecting spatial features or compressing the latent state reduces the cost of each rollout~\cite{costgrad,ddpwm,compact}. As CEM narrows its search, many candidates differ only slightly in their commanded motion. Dense CEM still evaluates each through a separate rollout.

These candidates share an observation history and hence an action-to-feature map. If the predicted features vary approximately linearly with small changes in the commands, a few rollouts can describe that variation across nearby candidates. \emph{ProbeShare} measures these local responses using selected action sequences, termed probes. It weights nearby probes using action geometry and a scene-conditioned similarity learned offline from the frozen model's predictions. When the weighted probe motion differs from a candidate's motion along a probed direction, shifting weight between opposite probes corrects the difference. The planner applies the adjusted weights to the full probe grids and compares the reconstructed features with the goal.

In later rounds, CEM may favor motions beyond the coverage of the initial probes. ProbeShare groups these promising sequences by their commanded motion. It evaluates a representative when its distance to the nearest probe is large relative to the spread of motions in its group. This single rollout supplies a feature grid that is reused to rescore nearby candidates in the current and subsequent rounds. After the robot executes the first command, the planner builds a fresh probe bank from the new observation.

This reuse enables a larger search under the same time budget. On RoboCasa's obstacle-reaching task with a five-step horizon, ProbeShare scores 40 times as many candidate sequences per episode as dense CEM at a target budget of 569\,s per episode. Across four paired starts, the end effector moves an average of 26.0\,cm closer to its goal over eighteen actions with ProbeShare and 17.0\,cm with dense CEM (Figure~\ref{fig:obstacle-budget}). Our contributions are as follows:
\begin{itemize}
  \item \emph{Feature sharing across action candidates.} We reconstruct each candidate's full feature grid from a few probe rollouts using action geometry, scene-conditioned similarity and a correction along the probed directions.
  \item \emph{Search-guided probe selection.} We use provisional goal scores and distances to existing probes to select representative actions whose predictions can be reused across nearby candidates and search rounds.
  \item \emph{Efficient closed-loop planning.} Across 16 paired starts on four RoboCasa tasks, ProbeShare accelerates episode planning by \mbox{$43.8$--$128.9\times$} over dense CEM at the same search size (Table~\ref{tab:closedloop}). At comparable planning times, mean terminal reward on obstacle reaching rises from 0.361 with dense CEM to 0.532 with ProbeShare.
\end{itemize}
\section{Related Work}
\paragraph{Joint-embedding world models for planning.}
Pretrained visual representations can encode observations for robot policies~\cite{r3m,vc1} and provide rewards for progress toward an image goal~\cite{vip,liv}. Planning in that space also requires predicting how actions change the features. DINO-WM~\cite{dinowm} learns these dynamics over patch features from a frozen DINOv2 encoder~\cite{dinov2}. Joint-embedding models learn visual features by predicting held-out image or video representations~\cite{ijepa,vjepa,murlabadia2026vjepa2_1}. V-JEPA~2 adds action-conditioned prediction to a pretrained video representation~\cite{vjepa2}, and RoboJEPA extends this approach across robot embodiments~\cite{robojepa}. This allows the planner to score predicted futures directly in the goal's feature space.

\paragraph{Efficient latent world-model planning.}
Rollout cost depends on the predicted representation and planning horizon. PlaNet evaluates sequences using latent dynamics learned from pixels~\cite{planet}. TD-MPC and TD-MPC2 use a learned terminal value to score short rollouts~\cite{tdmpc,tdmpc2}. For image goals, LeWorldModel learns compact JEPA dynamics from pixels~\cite{lewm}, and CompACT compresses observations into discrete tokens~\cite{compact}. Other methods allocate computation across space. CostGrad selects positions by goal sensitivity~\cite{costgrad}, Sparse Imagination predicts a sampled subset~\cite{sparseimagination}, and DDP-WM uses a lightweight update for the background~\cite{ddpwm}. Temporal methods can parallelize action-prefix predictions~\cite{fastlewm} or search at multiple time scales~\cite{hierarchicalwm,hjepa}. The cost of a planning decision also depends on the number of candidate sequences that require independent rollouts.

\paragraph{Interpolation across candidate actions.}
A single rollout can inform multiple candidates when their action sequences induce similar predicted scene changes. Classical interpolation uses the geometry of nearby samples to estimate a value between them~\cite{shepard,mls}. A learned representation can also align distances between inputs with similarities in their responses~\cite{deepkernel}. For RoboJEPA, each such response is a future feature grid. ProbeShare combines these grids using action geometry and a scene-conditioned measure of action similarity. Reconstructing a complete feature grid for each candidate allows the planner to retain the original goal-distance objective while changing how predictions are obtained.
\section{Method}
\begin{figure*}[t]
\centering\includegraphics[width=\linewidth]{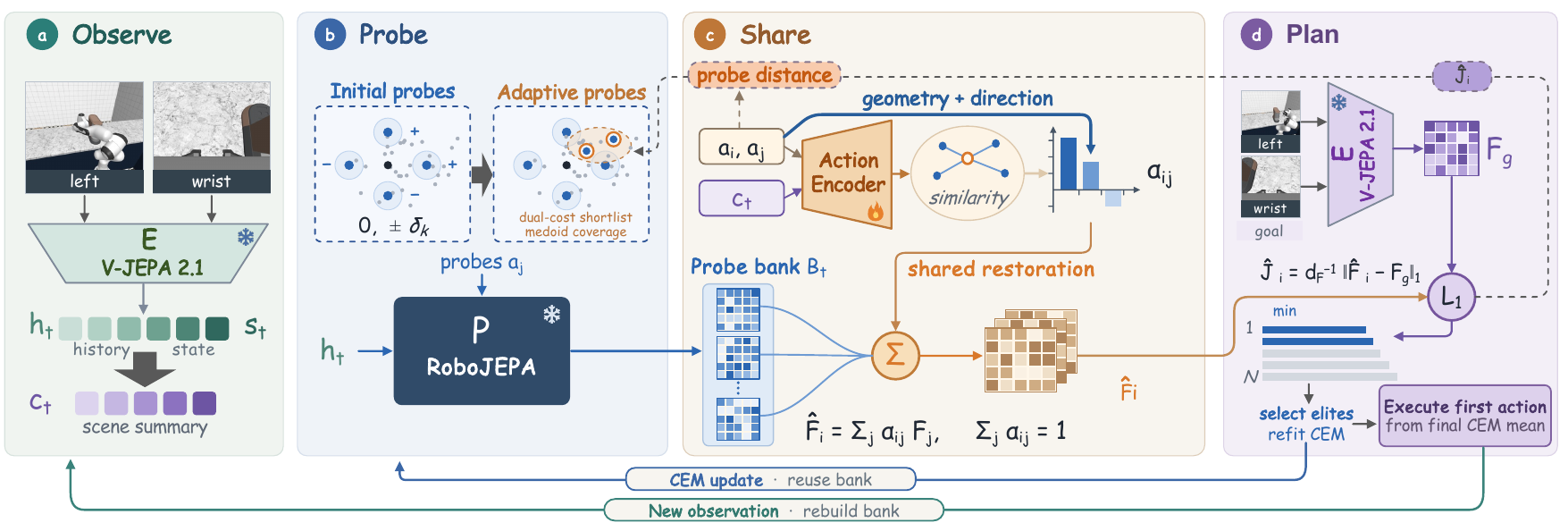}
\caption{One planning decision with ProbeShare. (a,b) Given the two-view observation history and robot state, frozen RoboJEPA predicts future feature grids for a few probe sequences. (c) Scene-conditioned coefficients combine these grids for each candidate. (d) CEM ranks candidates by mean L1 distance to the goal grid. Promising motions far from existing probes can trigger additional predictions. Search rounds share the probe bank until the robot executes the first command and observes again.}\label{fig:method}
\end{figure*}

At each planning decision, the observation history fixes a single map from action sequences to future feature grids. ProbeShare evaluates this map at a few probe sequences and stores their grids in a bank (Figure~\ref{fig:method}). To score a candidate, the planner combines these grids with weights specific to that motion. When the search favors a group of motions far from existing probes, a single new world-model evaluation supplies a grid for rescoring the group.

\subsection{Task and shared probe bank}
\label{sec:probe-bank}
At time $t$, the robot receives two camera views and its proprioceptive state $s_t$. A frozen V-JEPA 2.1 ViT-G/384 encoder~\cite{murlabadia2026vjepa2_1} produces the current feature grid $X_t\in\mathbb R^{600\times1664}$ and the goal grid $F_g$. Recent grids and robot states form the history $h_t$. For an $H$-step sequence $a=(u_0,\ldots,u_{H-1})$, each command contains translation, rotation and gripper coordinates. We use RoboJEPA's frozen 1B-parameter action-conditioned prediction head~\cite{robojepa} to compute the terminal grid $F_t(a)=P(h_t,a)$. The planning objective is
\begin{equation}
 a_t^\star\in\arg\min_{a\in\mathcal A_H}J_t(a),\qquad
 J_t(a)=\frac{\|F_t(a)-F_g\|_1}{d_F},
 \label{eq:dense-objective}
\end{equation}
where $d_F$ is the number of feature coordinates and $\mathcal A_H$ is the feasible action set. With $N$ sequences per round and $K$ rounds, dense CEM evaluates $NK$ candidate scores. ProbeShare estimates these scores from a shared bank of feature grids.

We initialize the bank $\mathcal B_t=\{(a_j,F_j)\}$ with a zero-command sequence and pairs of probes in opposite directions. The zero command provides a reference future $F_0$ for the current observation. Each pair measures the effect of reversing a perturbation along one active action coordinate. Cosine modes specify how that perturbation varies over the $H$ time steps, with amplitudes set by the initial candidate population. Evaluating each distinct probe gives $F_j=P(h_t,a_j)$. All probes reuse the observed-history key--value (KV) cache and maintain independent caches for their imagined futures.

Two sequences can yield the same predictor input. We identify such sequences using $v_t(a)$, which concatenates the action and state inputs supplied to the predictor, including action repetition. Equivalent inputs reuse a measured grid. Each original candidate still receives its score, preserving its multiplicity in CEM.

\subsection{Scene-conditioned action coefficients}
\label{sec:response-sharing}
Nearby commands provide useful interpolation points, but their effects also depend on the observed scene. To assign weights to neighboring probes for each new input, we summarize the scene by
\begin{equation}
 c_t=\operatorname{concat}\!\left(
 \operatorname{Pool}(X_t),\operatorname{Pool}(X_t-X_{t-1}),s_t\right),
 \label{eq:scene-summary}
\end{equation}
where each camera view is pooled separately. A context network conditions the action encoder on this summary. The encoder incorporates command values, cumulative motion and the prediction horizon. Projecting its output yields $z_i=e_\theta(a_i,c_t)W_\theta$, the representation used to compare actions in the current scene.

\paragraph{Local similarity.}
We first assign geometric weights $g_{ij}$ to nearby probes in predictor-input space. These weights combine inverse-squared-distance weighting~\cite{shepard} with a local convex reconstruction of the candidate input. The squared distance $d_{ij}^2=\|z_i-z_j\|_2^2$ in the learned representation then reweights those probes
\begin{align}
 \tau_i&=\max\!\left(\sum_j g_{ij}d_{ij}^2,\epsilon\right),\nonumber\\
 q_{ij}&=\frac{g_{ij}\exp[-\min(d_{ij}^2/\tau_i,\kappa)]}
 {\sum_\ell g_{i\ell}\exp[-\min(d_{i\ell}^2/\tau_i,\kappa)]}.
 \label{eq:local-weights}
\end{align}
Here $\epsilon$ and $\kappa$ stabilize the computation. Geometry selects the neighbors, and learned distance adjusts their relative influence. The resulting weights are nonnegative and sum to one.

\paragraph{Directional adjustment.}
The normalized weights determine the probes' relative contributions, but their weighted motion can still fall short of the candidate along a measured direction. Transferring weight from the negative probe to its positive partner changes that displacement while preserving the sum of the weights. Let row $k$ of $D$ be half the difference between the predictor inputs of the positive and negative probes in temporal direction $k$. For the input residual $r_i$, we compute
\begin{equation}
 r_i=v_t(a_i)-\sum_jq_{ij}v_t(a_j),\qquad
 \delta_i=\tfrac12 r_iD^\dagger,
 \label{eq:direction-coordinates}
\end{equation}
where $D^\dagger$ is the pseudoinverse. Starting from $q_i$, we add $\delta_{ik}$ to the positive probe's coefficient and subtract it from the negative probe's coefficient. We skip pairs with identical native inputs. The adjusted coefficients $\alpha_{ij}$ satisfy $\sum_j\alpha_{ij}=1$ and may have either sign.

Applying the same coefficients to the complete probe feature grids yields the candidate prediction
\begin{equation}
 \widehat F_i=\sum_j\alpha_{ij}F_j
 =F_0+\sum_j\alpha_{ij}(F_j-F_0).
 \label{eq:shared-features}
\end{equation}
The centered form identifies the shared prediction components. The reference $F_0$ is the zero-command future, and each $F_j-F_0$ is a measured change from that future. Within a given bank, only the coefficients vary across candidates.

\paragraph{Affine response reproduction.}
The paired adjustment reproduces the part of the candidate motion spanned by the probe directions. Writing $v_i=v_t(a_i)$ and $\Pi_D=D^\dagger D$, the remaining input mismatch is
\begin{equation}
 v_i-\sum_j\alpha_{ij}v_j=r_i(I-\Pi_D).
 \label{eq:input-reproduction}
\end{equation}
If $r_i$ lies in the retained directional subspace, the mismatch vanishes. When the feature response is affine over the candidate and contributing probes, matching the input in this way also recovers its full feature grid. For nonlinear responses, the derivation in the supplementary material relates reconstruction error to the probes' departures from the local affine map and the weights they receive.

\subsection{Adaptive probing and action search}
\label{sec:probe-reuse}
We score the reconstructed grids using the same mean L1 goal distance as the dense predictions,
\begin{equation}
 \widehat J_i=d_F^{-1}\|\widehat F_i-F_g\|_1.
 \label{eq:shared-score}
\end{equation}
Feature reconstruction retains signed differences from the goal until scoring; combining scalar probe costs discards this information. With candidates, probes and weights fixed, scoring reconstructed features more closely matches dense CEM's candidate selection on Obstacle and Push (see supplementary material).

For any goal grid, the reverse triangle inequality bounds the score error in terms of the feature reconstruction error,
\begin{equation}
 |\widehat J_i-J_t(a_i)|\leq d_F^{-1}
 \|\widehat F_i-F_t(a_i)\|_1\equiv\eta_i.
 \label{eq:cost-stability}
\end{equation}
Here $\eta_i$ measures terminal reconstruction error for the same candidate. Two candidates retain their dense ordering when their dense costs differ by more than $\eta_i+\eta_j$.

As CEM concentrates its samples, it may favor a cluster of motions poorly represented by the current bank. We shortlist promising candidates using two scores. The first scores the reconstructed grid in Eq.~\eqref{eq:shared-score}. The second, $\widetilde J_i=\sum_j\alpha_{ij}J_t(a_j)$, combines the probe costs directly. Taking absolute feature differences before or after combination can yield different rankings, so each score defines a separate shortlist. Within each shortlist, we group candidates by their cumulative command trajectories. We obtain a prediction for the group's medoid if its squared distance to the nearest probe exceeds the mean squared distance from group members to that medoid.

After adding the new grids to the bank, we rescore the current population. CEM selects its elites using the full-grid score $\widehat J_i$ and updates its sampling distribution. Subsequent rounds reuse the same bank. After the final round, the robot executes the first command of the mean sequence. The next observation defines a new action-to-feature map, so the planner starts a fresh bank. The supplementary material gives the search settings and blockwise scoring procedure.

\subsection{Learning action relationships}
We train the response head to model similarities in action effects within a scene, while keeping the visual encoder and world-model predictor frozen. The predictor supplies teacher responses on separate training and calibration trajectories from DROID and RoboCasa~\cite{droid,robocasa}. Each window contains a zero-command prediction, probe responses and additional query actions. Random queries cover a broad range of motions, while queries near promising candidates emphasize the finer distinctions relevant to search.

A differentiable kernel readout trains the action representation to predict query responses from support probes. We express flattened teacher responses relative to $F_0$ in a low-rank basis $B$ and collect the support coordinates in $C_S$. Let $K_{SS}$ and $K_{QS}$ be the support--support and query--support matrices of a kernel over projected action embeddings, conditioned on the zero-command reference. With a learned ridge parameter $\lambda>0$~\cite{ridge,gpml}, the readout predicts
\begin{equation}
 \begin{aligned}
 \widehat C_Q&=K_{QS}(K_{SS}+\lambda I)^{-1}C_S,\\
 \widehat F_Q&=F_0+\widehat C_QB.
 \end{aligned}
 \label{eq:training-readout}
\end{equation}
We train on the reconstructed responses and their goal costs using the objective
\begin{equation}
 \mathcal L=\lambda_{\rm f}\mathcal L_{\rm response}
 +\lambda_{\rm r}\mathcal L_{\rm random}
 +\lambda_{\rm c}\mathcal L_{\rm focused}.
 \label{eq:response-training}
\end{equation}
The response term supervises both feature values and differences between action responses. The losses for random and focused queries combine normalized cost error, correlation and pairwise ordering. Training uses sampled feature coordinates and reference-centered cost estimates. Checkpoint selection uses full-grid goal costs from the same kernel readout. The supplementary material gives the loss weights, data splits and optimization details.

Training and planning share the learned action distance. The differentiable kernel readout trains the action encoder to map motions with similar feature responses to nearby representations. During planning, this distance reweights the geometric neighbors in the current bank (Eq.~\eqref{eq:local-weights}), and the paired probes correct the motion along their measured directions (Eq.~\eqref{eq:direction-coordinates}). Offline training supplies the action metric; online rollouts supply the full feature grids combined during planning. With the encoder fixed, a new probe can therefore change nearby candidates' scores as soon as its grid enters the bank. Table~\ref{tab:coefficient-ablation} compares estimated and dense goal costs on fixed candidate pools and reports closed-loop reward on Obstacle.
\section{Experiments}
\begin{table*}[t]\centering\small\figuretype
\caption{Closed-loop planning on A800 with paired starts. Reach and Object Reach report mean progress over five actions. Obstacle reports progress after eighteen actions. Push counts episodes ending closer to the goal. Next-step L1 averages per-position error on executed actions. Time is the mean spent planning per episode. Parentheses give speedup over dense RoboJEPA. \textbf{Bold} marks the best displayed value and \underline{underlining} the next distinct value.}\label{tab:closedloop}
\setlength{\tabcolsep}{1.15pt}
\renewcommand{\arraystretch}{0.94}
\newcommand{\speedupnote}[1]{{\footnotesize\mdseries\upshape(#1$\times$)}}
\newcommand{\methodhead}[3]{\multicolumn{1}{c}{\shortstack{\strut #1\,{\footnotesize\cite{#3}}\\{\footnotesize(#2)}}}}
\newlength{\benchmarkwidth}
\settowidth{\benchmarkwidth}{25,728.4}
\begin{tabular*}{\linewidth}{@{\extracolsep{\fill}}lc*{5}{w{l}{\benchmarkwidth}}l@{\extracolsep{0pt}\hspace{0.28em}}l@{}}\toprule
 & & & \multicolumn{4}{c}{Acceleration baselines adapted to RoboJEPA} & \multicolumn{2}{c@{}}{} \\
\cmidrule(lr){4-7}
Task & Metric & \methodhead{RoboJEPA}{arXiv 2026}{robojepa} & \methodhead{CostGrad}{NeurIPS 2026}{costgrad} & \methodhead{DDP-WM}{ICML 2026}{ddpwm} & \methodhead{CompACT}{CVPR 2026}{compact} & \methodhead{Sparse Imagination}{ICLR 2026}{sparseimagination} & \multicolumn{2}{c@{}}{\shortstack{\strut\textbf{ProbeShare}\\{\footnotesize\textbf{(Ours)}}}}\\\midrule
Reach & Progress reward $\uparrow$ & \textbf{0.935} & 0.853 & 0.908 & 0.046 & \underline{0.921} & 0.863 & \\
 & Next-step L1 $\downarrow$ & \textbf{0.145} & 0.261 & 0.166 & 0.370 & \underline{0.159} & 0.172 & \\
 & Plan time (s) $\downarrow$ & 1,402.1 & 1,080.0 & 840.0 & \underline{554.9} & 978.8 & \textbf{14.4} & \speedupnote{97.4}\\
\midrule[0.3pt]
Object Reach & Progress reward $\uparrow$ & 0.336 & 0.362 & \underline{0.620} & 0.013 & 0.590 & \textbf{0.657} & \\
 & Next-step L1 $\downarrow$ & 0.224 & 0.309 & \underline{0.218} & 0.394 & 0.221 & \textbf{0.205} & \\
 & Plan time (s) $\downarrow$ & 1,390.3 & 827.6 & 793.1 & \underline{408.1} & 848.4 & \textbf{10.8} & \speedupnote{128.9}\\
\midrule[0.3pt]
Obstacle & Terminal reward $\uparrow$ & \underline{0.577} & \textbf{0.782} & 0.037 & 0.224 & 0.331 & 0.532 & \\
 & Next-step L1 $\downarrow$ & \textbf{0.202} & 0.312 & 0.210 & 0.352 & 0.217 & \underline{0.204} & \\
 & Plan time (s) $\downarrow$ & 24,941.7 & 12,352.9 & 12,551.5 & \underline{8,424.0} & 16,834.5 & \textbf{569.0} & \speedupnote{43.8}\\
\midrule[0.3pt]
Push & Ends closer $\uparrow$ & \textbf{4/4} & \textbf{4/4} & \textbf{4/4} & \underline{3/4} & \textbf{4/4} & \textbf{4/4} & \\
 & Next-step L1 $\downarrow$ & \underline{0.139} & 0.175 & \underline{0.139} & 0.390 & 0.143 & \textbf{0.133} & \\
 & Plan time (s) $\downarrow$ & 25,728.4 & 14,574.8 & 12,540.9 & \underline{7,589.2} & 13,066.3 & \textbf{535.4} & \speedupnote{48.1}\\
\bottomrule\end{tabular*}\end{table*}
At a fixed search size, we quantify the planning time saved by shared predictions. At comparable planning times, we evaluate the robot's progress toward its goal under each planner. Probe ablations and fixed-input timings then relate these outcomes to the construction and reuse of the probe bank.

\subsection{Experimental Setup}
\label{sec:experimental-setup}
\paragraph{Tasks and outcomes.}
We evaluate each of four RoboCasa~\cite{robocasa} tasks in two kitchens with two paired starts per kitchen. We follow RoboJEPA's task metrics~\cite{robojepa}. Reach and Object Reach use one-step planning and average normalized progress $r_t=\max(0,1-d_t/d_0)$ over five actions. After action $t$, we measure the distance $d_t$ to the target from the end effector on Reach and from the held mug on Object Reach. Obstacle and Push execute eighteen actions with horizons up to five. Obstacle scores terminal progress $r_{18}$. On Push, an episode counts if the object ends closer to its target ($d_{18}<d_0$)~\cite[Sec.~3.3, Fig.~15]{robojepa}. We report this count only for benchmark comparison and assess closed-loop planning effectiveness from goal progress on the reaching tasks.

\paragraph{Baselines.}
Dense CEM runs the frozen RoboJEPA predictor for every candidate. Four adaptations of the same backbone reduce the computation required for each prediction. CostGrad~\cite{costgrad} selects 300 of 600 positions by goal sensitivity, and Sparse Imagination~\cite{sparseimagination} uses trained adapters on a random subset of the same size. DDP-WM~\cite{ddpwm} predicts selected features and learns a cheaper update for the remaining features. CompACT~\cite{compact} predicts in a compact discrete space and decodes a full grid. The supplementary material gives the adaptations and training splits.

\paragraph{Search and timing.}
In Table~\ref{tab:closedloop}, paired planners share the initial observations, state, goal and candidate population. Each round scores 1,024 candidates and retains twenty elites. The first decision uses fifteen rounds and later decisions use ten. Episode planning time sums all search times from sampling through the final distribution update, including prediction, reconstruction and scoring. Speedup compares mean episode planning times.

On A800, all planners share the observed-history KV cache and keep candidate futures separate. Dense CEM uses compiled batches of sixteen and the four adaptations use compiled batches of eight. ProbeShare uses batches of eight, compiled on short tasks and eager on long tasks. Fixed-input comparisons use compiled batches of eight for both planners.

\paragraph{Prediction error.}
Before execution, we record the predicted features for the robot's next observation. For a multi-step plan, ProbeShare applies the planning coefficients to the first-step probe features. After execution, we encode the two camera views and measure mean absolute error over the predicted positions and channels. Table~\ref{tab:closedloop} reports this error along each planner's own trajectory. For a paired comparison, the dense predictor also forecasts the next observation from ProbeShare's history and executed command.

\subsection{Closed-Loop Planning Performance}
\paragraph{Planning time and goal progress.}
At 1,024 candidates per round, ProbeShare achieves the shortest planning time of the six planners in Table~\ref{tab:closedloop} on each of the 16 paired starts (see supplementary material). On Object Reach, mean episode planning time falls from 1,390.3\,s with dense CEM to 10.8\,s with ProbeShare, a $128.9\times$ speedup. Mean progress on these four starts is 0.336 with dense CEM and 0.657 with ProbeShare. The speedups over dense CEM are $97.4\times$ on Reach, $43.8\times$ on Obstacle and $48.1\times$ on Push. Across the four tasks, ProbeShare plans $14.2$--$38.6\times$ faster than CompACT, the fastest within-rollout adaptation.

\begin{figure}[t]
\centering
\includegraphics[width=\columnwidth]{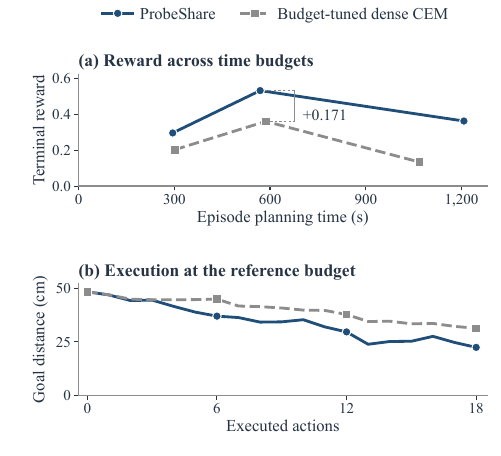}
\caption{Obstacle reaching at comparable planning times. (a) ProbeShare earns higher mean terminal reward across four paired starts at each of three target budgets. (b) Mean end-effector distance to the goal after each action at the reference budget, where ProbeShare scores $40\times$ as many candidates as dense CEM.}\label{fig:obstacle-budget}
\end{figure}

\paragraph{Goal progress at matched planning time.}
To assess how computational savings affect control, we compare goal progress at comparable episode planning times. For each of three Obstacle time budgets, we configure each planner's population and number of search rounds. Both retain twenty elites per round (see supplementary material). ProbeShare achieves higher mean terminal reward at all three budgets (Figure~\ref{fig:obstacle-budget}a). At the reference budget, it scores 189,440 sequences in 569.0\,s using about 1,051 full rollouts. Dense CEM scores 4,736 sequences in 586.8\,s. Over eighteen actions, the arm reduces its distance to the goal by an average of 26.0\,cm with ProbeShare and 17.0\,cm with dense CEM (Figure~\ref{fig:obstacle-budget}b). The corresponding terminal rewards are 0.532 and 0.361.

\paragraph{Progress on Push.}
Dense CEM with the frozen 1B RoboJEPA predictor reduces object-to-goal distance by 0.35\,mm on average versus 0.40\,mm for ProbeShare on the same four starts. Both score $4/4$. The small advance with dense CEM is consistent with smaller RoboJEPA models' difficulty predicting gripper--object contact~\cite[Sec.~3.5]{robojepa}.

\paragraph{Predicting the next observation.}
The planners generate different trajectories, so we also compare their predictions on the same executed actions. For the 184 actions that ProbeShare executes in 16 episodes, both predictors receive the same history and command. We compare their forecasts with the next observation at all 600 feature positions. On each of the four tasks, ProbeShare's mean next-step L1 is within $1\%$ of the dense predictor's (see supplementary material).

\subsection{Ablations of Probing and Weighting}
Adding probes tests the value of new predictions with the weighting rule fixed. Keeping the bank fixed tests how the weighting rule combines those predictions.

\begin{table*}[t]\centering\small\figuretype
\caption{Effect of adding probes during search. Both variants start with the same directional probes. \textbf{Full} can add probes when promising candidates are far from existing probes. Task scores follow Table~\ref{tab:closedloop}. Probe counts and planning time are per-episode means. Next-step L1 measures prediction error on each variant's executed actions.}\label{tab:probe-ablation}
\setlength{\tabcolsep}{3pt}
\begin{tabular*}{\linewidth}{@{\extracolsep{\fill}}lcccccccc@{}}\toprule
 & \multicolumn{6}{c}{Closed-loop planning} & \multicolumn{2}{c}{Next-step prediction}\\
\cmidrule(lr){2-7}\cmidrule(l){8-9}
 & \multicolumn{2}{c}{Task score $\uparrow$} & \multicolumn{2}{c}{Probe rollouts} & \multicolumn{2}{c}{Plan time (s) $\downarrow$} & \multicolumn{2}{c}{L1 $\downarrow$}\\
\cmidrule(lr){2-3}\cmidrule(lr){4-5}\cmidrule(lr){6-7}\cmidrule(l){8-9}
Task & Init only & \textbf{Full} & Init only & \textbf{Full} & Init only & \textbf{Full} & Init only & \textbf{Full}\\\midrule
Reach & 0.645 & \textbf{0.863} & 35.0 & 182.5 & 7.1 & 14.4 & 0.188 & \textbf{0.172}\\
Object Reach & 0.453 & \textbf{0.657} & 35.0 & 96.2 & 6.6 & 10.8 & \textbf{0.203} & 0.205\\
Obstacle & 0.119 & \textbf{0.532} & 498.0 & 1,051.2 & 212.4 & 569.0 & 0.207 & \textbf{0.204}\\
Push & 4/4 & 4/4 & 498.0 & 920.2 & 197.1 & 535.4 & 0.198 & \textbf{0.133}\\
\bottomrule\end{tabular*}\end{table*}
\begin{figure}[b]
\centering
\includegraphics[width=\columnwidth]{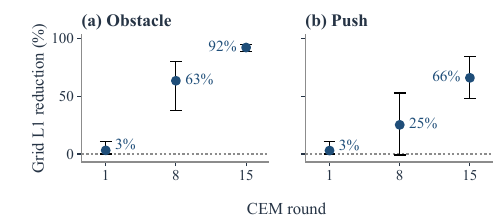}
\caption{Five-step terminal grid L1 against dense RoboJEPA on one input per task. For each round's unprobed candidates, we express the paired reductions from adding probes as a percentage of their mean error with the initial bank. Dots show means and bars the 10th--90th percentiles.}\label{fig:terminal-feature-gain}
\end{figure}
\begin{figure*}[t]
\begin{minipage}[t]{\dimexpr(\textwidth-\columnsep)/2\relax}
  \captionsetup{type=figure}
\centering
\includegraphics[width=\linewidth]{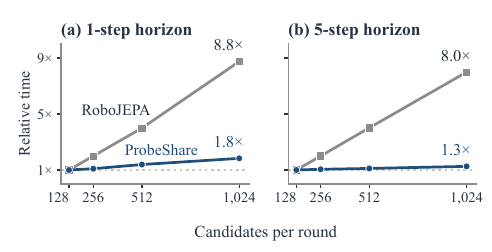}
\caption{With eight times as many candidates, ProbeShare takes $1.8\times$ its original planning time at $H=1$ and $1.3\times$ at $H=5$. Each curve is normalized to its own 128-candidate time on A800. Points average the three-run medians for the two tasks at each horizon.}\label{fig:candidate-scaling}
\end{minipage}\hfill
\begin{minipage}[t]{\dimexpr(\textwidth-\columnsep)/2\relax}
  \captionsetup{type=figure}
\centering
\includegraphics[width=\linewidth]{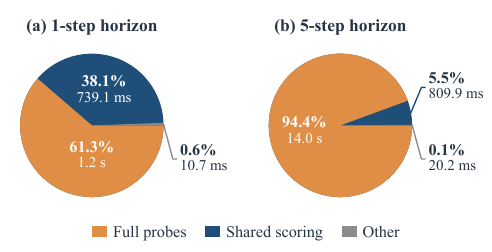}
\caption{Full probes account for $61.3\%$ of ProbeShare's GPU kernel time at $H=1$ and $94.4\%$ at $H=5$. The totals are 1.9 and 14.8\,s on A800 at 1,024 candidates per round. Shared scoring includes coefficients and feature restoration.}\label{fig:planning-profile}
\end{minipage}
\end{figure*}
\begin{figure*}[t]\centering
\includegraphics[width=\linewidth]{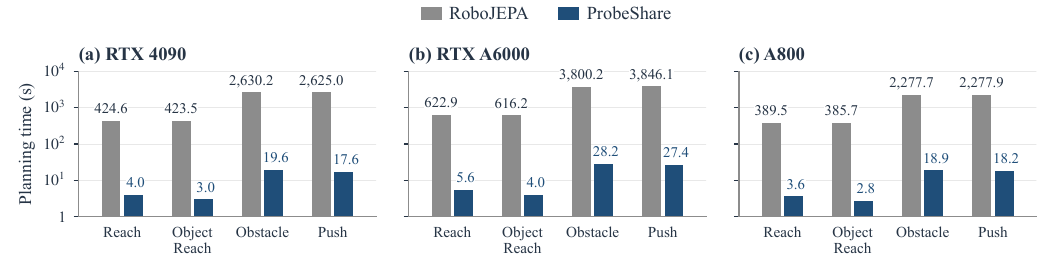}
\par
\begingroup\footnotesize\figuretype
\setlength{\tabcolsep}{10pt}
\renewcommand{\arraystretch}{1.0}
\begin{tabular*}{\linewidth}{@{\extracolsep{\fill}}lcccc@{}}\toprule
Speedup ($\times$) & Reach & Object Reach & Obstacle & Push\\\midrule
RTX 4090 & 105.4 & 140.1 & 134.0 & 149.0\\
RTX A6000 & 111.7 & 152.9 & 134.8 & 140.5\\
A800 & 107.3 & 139.0 & 120.4 & 125.5\\
\bottomrule\end{tabular*}
\endgroup
\caption{Across twelve task--device pairs, speedups for one complete search range from $105.4\times$ to $152.9\times$. Bars give median seconds on a shared log scale. Each timing pair uses the same input and GPU, three timed runs after compilation and warmup, 1,024 candidates, fifteen rounds and batches of eight. Reach and Object Reach use $H=1$, and Obstacle and Push use $H=5$.}\label{fig:hardware-speedup}
\end{figure*}

\paragraph{Where added probes help.}
In 53 of 56 decisions across Reach, Obstacle and Push, an added probe is closer in action space to the selected plan than any initial probe. We compare initial and expanded banks on identical unprobed candidates from rounds 1, 8 and 15 of five-step Obstacle and Push searches (Figure~\ref{fig:terminal-feature-gain}). Adding probes reduces mean terminal feature L1 in all six comparisons; round 15 reductions are $92.0\%$ on Obstacle and $65.8\%$ on Push.

\begin{table}[t]\centering\small\figuretype
\caption{Action weighting in five-step planning. Candidate scoring uses the same probes and actions in six pools. The closed-loop comparison uses the same search schedule on four Obstacle starts.}\label{tab:coefficient-ablation}
\setlength{\tabcolsep}{2.5pt}
\begin{tabular*}{\columnwidth}{@{\extracolsep{\fill}}lcc@{}}\toprule
\multicolumn{3}{@{}l}{\emph{Candidate scoring against the dense predictor}}\\
Coefficients & Cost MAE $\downarrow$ & Spearman $\rho$ $\uparrow$\\\midrule
Without learned similarity & \textbf{0.0218} & 0.541\\
Without directional adjustment & 0.0245 & \underline{0.584}\\
\textbf{ProbeShare} & \underline{0.0219} & \textbf{0.615}\\
\midrule
\multicolumn{3}{@{}l}{\emph{Closed-loop obstacle reaching}}\\
Coefficients & Reward $\uparrow$ & Plan time (s)\\\cmidrule(lr){1-3}
Without learned similarity & 0.351 & 569.5\\
\textbf{ProbeShare} & \textbf{0.532} & 569.0\\
\bottomrule\end{tabular*}\end{table}
\paragraph{Progress under repeated planning.}
To assess how these measurements affect control, we run the planner with the initial bank alone and with probe additions, keeping the trained encoder and search schedule fixed (Table~\ref{tab:probe-ablation}). Over eighteen actions on the four Obstacle starts, the arm moves an average of 5.5\,cm closer to its goal with the initial bank and 26.0\,cm with additions. Reward also increases on all four Reach starts.

On the four Object Reach starts, mean progress after the first action is 0.537 with added probes and 0.192 with the initial bank. The corresponding five-action means are 0.657 and 0.453. On Push, next-step feature L1 falls on all four starts with added probes, from 0.198 to 0.133 on average along the planners' own trajectories.

\paragraph{How candidates weight the probes.}
To separate coefficient quality from probe selection, we keep the probes fixed. Learned similarity raises the Spearman correlation between estimated and dense goal costs for unprobed candidates in five of six pools (Table~\ref{tab:coefficient-ablation}). Adding directional adjustment to learned similarity reduces goal-cost MAE from 0.0245 to 0.0219. In a separate closed-loop comparison on Obstacle, learned similarity raises mean reward from 0.351 to 0.532 at about 569\,s per episode under the same probe selection rule and search schedule.

\subsection{Computational Analysis}
\label{sec:computation-scaling}
\paragraph{Full predictions as the search grows.}
To measure how the number of full rollouts scales with search size, we vary $N\in\{128,256,512,1024\}$ on one fixed input per task. Each timing pair uses the same A800, fifteen rounds, twenty elites and compiled batches of eight. Increasing $N$ from 128 to 1,024 adds 13,440 candidate scores per search. Dense CEM requires one more full rollout for each added score. ProbeShare requires eight additional full rollouts on average at $H=1$ and five more at $H=5$. At $H=5$ and $N=1024$, it scores all 15,360 candidates with 68 full rollouts on average, a $99.6\%$ reduction. Both planners still score the requested population in every round. The reduction in rollouts therefore results from sharing predictions across the search.

This reuse also changes how planning time scales with $N$ (Figure~\ref{fig:candidate-scaling}). An eightfold increase in candidates multiplies dense planning time by 8.8 at $H=1$ and 8.0 at $H=5$. ProbeShare's time grows by factors of 1.8 and 1.3, respectively. At $H=5$, its time rises from 14.7 to 18.5\,s as dense time rises from 284.9 to 2,277.8\,s. The speedup therefore grows from $19.3\times$ to $122.9\times$ at the same search budgets.

\paragraph{GPU time by planning horizon.}
GPU profiles at $N=1024$ separate the cost of obtaining probe grids from the cost of sharing them (Figure~\ref{fig:planning-profile}). Computing coefficients, reconstructing grids and scoring candidates together require 739.1\,ms on the one-step inputs and 809.9\,ms on the five-step inputs. Full probe rollouts account for 1.2\,s ($61.3\%$) of attributed kernel time at $H=1$ and 14.0\,s ($94.4\%$) at $H=5$. As the horizon grows, the predictor executes more steps per probe while shared scoring still reconstructs a terminal grid.

\subsection{Cross-Hardware Efficiency}
We repeat the fixed-input timing on RTX 4090, RTX A6000 and A800 with the same encoded observations, 1,024 candidates and compiled batches of eight. Speedup over dense CEM ranges from $105.4\times$ on Reach with the RTX 4090 to $152.9\times$ on Object Reach with the RTX A6000 (Figure~\ref{fig:hardware-speedup}).

These fixed-input measurements isolate search cost, whereas episode timings include replanning as observations change. ProbeShare accelerates planning in both settings.

\section{Conclusion}
The cost of evaluating a candidate action with a world model can be much lower than that of predicting its future independently. ProbeShare reconstructs candidate futures from measured responses to nearby actions and evaluates a new probe when search moves beyond their coverage. Across four RoboCasa tasks, ProbeShare reduces episode planning time by a factor of 43.8 to 128.9 relative to dense CEM at the same search size. In four paired obstacle-reaching trials, ProbeShare scores 40 times as many candidate sequences as dense CEM in comparable planning time. After eighteen actions, the arm is on average 8.9\,cm closer to its goal with ProbeShare.
\nocite{adamw}
{\small\bibliographystyle{ieeenat_fullname}\bibliography{main,related}}
\end{document}